\documentclass[runningheads]{llncs}
\usepackage{graphicx}
\usepackage{latexsym}
\usepackage{booktabs}
\usepackage{amsmath} 
\usepackage[linesnumbered,ruled]{algorithm2e}
\begin{document}
\title{Abstaining Classification When Error Costs are Unequal and Unknown}
%
%
\author{Hongjiao Guan\inst{1} \and
Yingtao Zhang\inst{1} \and
H. D. Cheng\inst{2} \and
Xianglong Tang\inst{1}}
\authorrunning{H. Guan et al.}
%
\institute{School of Computer Science and Technology, Harbin Institute of Technology, Harbin 150001, China \\
\email{guanhongjiao2008@163.com} \and
School of Computer Science, Utah State University, Logan, UT 84322, USA
}
\maketitle              
\begin{abstract}
Abstaining classification aims to reject to classify the easily misclassified examples, so it is an effective approach to increase the classification reliability and reduce the misclassification risk in the cost-sensitive applications. In such applications, different types of errors (false positive or false negative) usually have unequal costs. And the error costs, which depend on specific applications, are usually unknown. However, current abstaining classification methods either do not distinguish the error types, or they need the cost information of misclassification and rejection, which are realized in the framework of cost-sensitive learning. In this paper, we propose a bounded-abstention method with two constraints of reject rates (BA2), which performs abstaining classification when error costs are unequal and unknown. BA2 aims to obtain the optimal area under the ROC curve (AUC) by constraining the reject rates of the positive and negative classes respectively. Specifically, we construct the receiver operating characteristic (ROC) curve, and stepwise search the optimal reject thresholds from both ends of the curve, until the two constraints are satisfied. Experimental results show that BA2 obtains higher AUC and lower total cost than the state-of-the-art abstaining classification methods. Meanwhile, BA2 achieves controllable reject rates of the positive and negative classes.

\end{abstract}
\section{Introduction}
Pattern classification techniques have been widely applied to solve practical issues, such as face/text recognition, fault detection, medical diagnosis and so on. A large number of approaches have been proposed to increase the classification accuracy of total examples. However, they neglect the classification reliability of each individual example, especially, in the risk-associated fields. In such fields, the wrong classification of a specific example leads to serious consequences, such as enormous economic loss or irretrievable death. Abstaining classification~\cite{pietraszek2005optimizing} or classification with reject option~\cite{Kamiran2017Exploiting,Wang2017Fault} is helpful to improve the reliability and reduce the risk by abstaining the uncertain examples of low membership degree, since these examples are easily misclassified. 

There are two reject rules in abstaining classification: Chow's rule and ROC-based rule. Chow~\cite{chow1970optimum} initially presents the optimal reject rule in the framework of Bayesian theory. In Chow's rule, the classification of an example is rejected if its maximum posterior probability is less than a given threshold. Without distinction of error types and correct recognition types, the threshold is related to the ratio of $(w_r-w_c)/(w_e-w_c)$, where $w_r$, $w_c$ and $w_e$ are the costs of rejecting, correctly classifying or misclassifying an example. Chow's optimal reject rule is obtained only if the exact knowledge of the posterior probabilities is known, which is impossible in practice. The alternative way is to use estimated posterior probabilities, such as the probabilistic outputs in neural networks~\cite{giusti2002theoretical}. Also, the outputs of non-probabilistic classifiers can be converted into posterior probabilities using the sigmoid transformation function~\cite{platt1999probabilistic}, isotonic regression~\cite{zadrozny2002transforming}, the histogram method~\cite{li2006confidence}, or bootstrap sampling method~\cite{xie2006bootstrap}. However, the accuracy of the estimation significantly influences the performance of reject classification. Other methods, such as estimating the data distribution~\cite{devarakota2007confidence}, probability density function~\cite{ishidera2003confidence}, and confidence interval \cite{devarakota2008reliability}, are proposed for classification with rejection. 

ROC-based rule is proved to be theoretically equivalent with Chow's rule under the general cost term that distinguishes the wrong and correct recognition types~\cite{santos2005optimal}. And researches have shown that ROC-based rule performs better than Chow's rule on real world datasets \cite{xie2006bootstrap,marrocco2007empirical}. ROC-based rule is initially proposed in~\cite{tortorella2000optimal}, where two points on the ROC curve corresponding to two reject thresholds are determined by minimizing the total cost. This method is implemented using support vector machine (SVM) in~\cite{tortorella2004reducing}. Pietraszek~\cite{pietraszek2005optimizing} proposes a ROC-based bounded-abstention model (BA), which minimizes a cost-weighted function while keeping the overall reject rate below a given value. A twin SVM with reject option (RO-TWSVM) is proposed in \cite{lin2017twin}, which improves the previous SVM with reject option (RO-SVM) \cite{tortorella2004reducing} using twin SVM instead of SVM. These abstaining classification methods are cost-sensitive, which rely on cost information that is usually unknown in practical applications.

In this paper, we propose a ROC-based abstaining classification method, bounded-abstention with two constraints of reject rates (BA2), to overcome the limitations of using posterior probabilities and cost information. Note that in the paper we only consider the binary classification. We expect to maximize the area under the ROC curve (AUC) under two constraints that the reject rates of the positive class and the negative class do not exceed two given bounds, respectively. This is realized by stepwise searching two points on the ROC curve from two endpoints. When the two constraints are satisfied, the searching process stops, and the final two points are corresponding to the reject thresholds.

The proposed method has several advantages. BA2 is developed to obtain maximum AUC value rather than to minimize the total cost, which skillfully avoids setting the unknown cost term. Furthermore, BA2 distinguishes the error types and restrains the reject rates of the positive class and the negative class separately. This is beneficial to control the respective performance of two classes when the error costs are unequal.

\section{Proposed Method}
The ROC curve depicts different operating points using the false positive rates ($fpr$) as the $x$-axis and the true positive rates ($tpr$) as the $y$-axis, which can visualize the performance of a binary classifier. Note that $fpr=N_{fp}/N_{neg}$, i.e., the number of misclassified negative examples ($N_{fp}$) divided by the number of all negative examples ($N_{neg}$); $tpr=N_{tp}/N_{pos}$ i.e., the number of positive examples that are correctly classified ($N_{tp}$) divided by the number of all positive examples ($N_{pos}$). The ROC curve is insensitive to unequal misclassification costs~\cite{Fawcett06anintroduction}, which is an effective tool to analyze classifiers' behavior.

Assume that in the two-class classification problem, an example $x$ obtains its score $s(x)$ measuring the likeliness of belonging to positive class. We aim to obtain the two reject thresholds $t_1$ and $t_2$ ($t_1 < t_2$), and the corresponding reject rule is as follows: 
\begin{equation}\label{eq1}
class = \begin{cases}
positive,  & \text{if } s(x)>t_2;  \\
negative, & \text{if } s(x)<t_1;   \\
reject, & \text{else}.
\end{cases}
\end{equation}

The idea of the proposed BA2 method is to maximize AUC under two constraints of two classes' reject rates. The problem is formalized as:
\begin{equation} \label{e2}
\begin{split}
&\max_{t_1,t_2} \hspace{0.5em}AUC(t_1,t_2)  \\
&s.t.\hspace{0.5em} rnr \leq n_{max} \\
& \hspace{1.8em}rpr \leq p_{max}  
\end{split}
\end{equation}
The two constraints are $rnr \leq n_{max}$ and $rpr \leq p_{max}$. $rnr$ is defined as the ratio of the number of rejected negative examples ($N_{rn}$) divided by the number of all negative examples ($N_{neg}$); $rnr=N_{rn}/N_{neg}$. $rpr$ is defined as the ratio of the number of rejected positive examples ($N_{rp}$) divided by the number of all positive examples ($N_{pos}$); $rpr=N_{rp}/N_{pos}$. We use $n_{max}$ and $p_{max}$ to denote the maximum reject rates that $rnr$ and $rpr$ should not exceed, respectively. 

\begin{algorithm}
\caption{Constructing the bounded-abstention classifier with two constraints (BA2)} 
    \KwIn{the ROC curve described by multiple points ($fpr, tpr$); $n_{max}$ and $p_{max}$, the maximum reject rates that $rnr$ and $rpr$ should not exceed；\\}
    \KwOut{$t_1$ and $t_2$, the two reject thresholds\\}
    \BlankLine
    Initialization: $l \leftarrow 1, r \leftarrow 1$, $x_1 \leftarrow 1-0.01 \times r, x_2 \leftarrow 0.01 \times l$; \\
    $times \leftarrow max(p_{max},n_{max}) / min(p_{max},n_{max})$;\\
     \While {$x_1 > x_2$}{
     Compute $rnr$ and $rpr$ according to equations (\ref{e3}) and (\ref{e4}), respectively;\\
     \If{$rnr>n_{max}$ and $rpr>p_{max}$} {
     \eIf {$n_{max} > p_{max}$}
     {\eIf {$rnr > rpr \times times$}{$r \leftarrow r+1$; \\}{$l \leftarrow l+1$;\\}}
     {\eIf {$rpr > rnr \times times$}{$l \leftarrow l+1$; \\}{$r \leftarrow r+1$;\\}}
     $x_1 \leftarrow 1-0.01 \times r, x_2 \leftarrow 0.01 \times l$; \textbf{continue};\\
     }
     \If {$rnr>n_{max}$ and $rpr \leq p_{max}$}{$r \leftarrow r+1, x_1 \leftarrow 1-0.01 \times r$; \textbf{ continue};\\}
     \If {$rnr \leq n_{max}$ and $rpr > p_{max}$}{$l \leftarrow l+1, x_2 \leftarrow 0.01 \times l$; \textbf{ continue};\\}
     \textbf{break};\\
     Calculate the score thresholds corresponding to the locations of ($x_1,f(x_1)$) and $(x_2,f(x_2))$, i.e., $t_1$ and $t_2$.}
\end{algorithm}

We regard the ROC curve as a function $f$ in the two-dimensional ROC space, so we can denote $tpr=f(fpr)$. And we denote $fpr$ as $x$ and $tpr$ as $y$. ($x_1,f(x_1)$) and ($x_2,f(x_2)$) are the two points on the ROC curve ($x_1>x_2$). According to the definitions, $rnr$ and $rpr$ can be expressed as \cite{vanderlooy2009roc}:
\begin{align}
rnr &= x_1-x_2     \label{e3}\\
rpr &= f(x_1)-f(x_2) \label{e4}
\end{align}
Let $(x_1,f(x_1))$ and $(x_2,f(x_2))$ start from points (1, 1) and (0, 0), and step-wisely compute $rnr$ and $rpr$ with step 0.01, until $rnr \leq n_{max}$ and $rpr \leq p_{max}$. Algorithm 1 is the pseudo-code of constructing BA2. Note that, since the ROC curve obtained on real datasets has concavities, the convex hull of points in the ROC plane (ROCCH) is usually constructed \cite{fawcett2004roc}. In the paper, when we mention the ROC curve, it refers to the corresponding ROCCH curve.

Lines 5-26 in Algorithm 1 is the process of searching the two optimal points $(x_1,f(x_1))$ and $(x_2,f(x_2))$ on the ROC curve. When the constraints are not satisfied, we increase the value of $x_2$ or decrease the value of $x_1$ with step 0.01. Now we analyze the searching process according to whether $n_{max}$ and $p_{max}$ are equal. 

\begin{figure}[htbp]  
\centering
\includegraphics[width=0.9\textwidth]{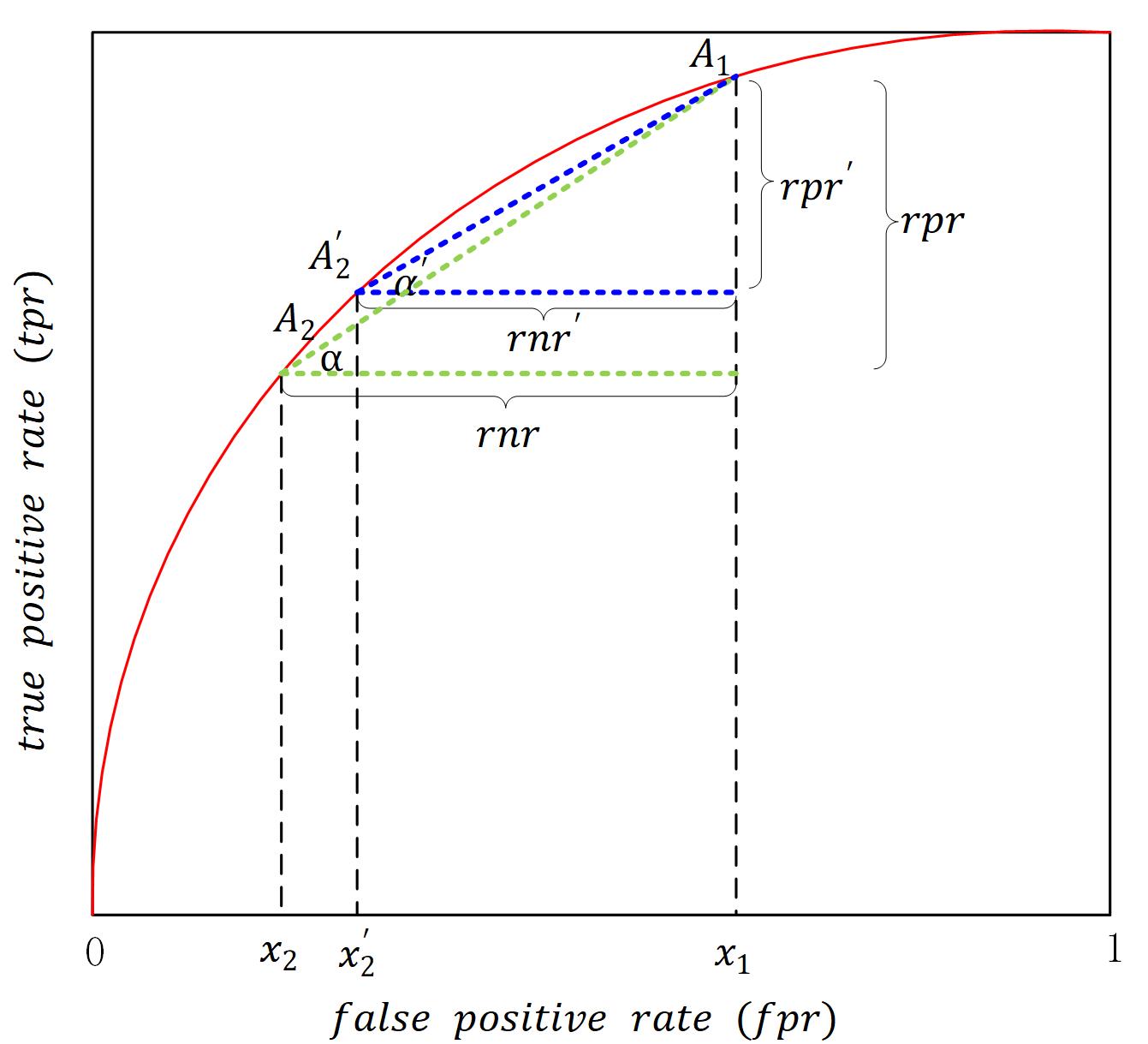}
\caption{The illustration of the searching process}\label{f0}
\end{figure}
When $n_{max} = p_{max}$, $times$ = 1 and lines 6-18 in Algorithm 1 can be simplified as:
\begin{equation}   \label{e5}
\begin{split} 
&\textbf{if } rnr>rpr \textbf{ then}\\
&\hspace{1em}r \leftarrow r+1;  \\
&\textbf{else} \\
&\hspace{1em}l \leftarrow l+1; \\
&\textbf{end} 
\end{split}
\end{equation}
This statement means that if the rejected negative rate is larger than the rejected positive rate, decrease $x_1$; otherwise, increase $x_2$. \cite{vanderlooy2009roc} indicates that if $rnr$ equals $rpr$, the AUC of the abstention ROC (the ROC curve obtained by the abstaining classifier)is always larger than that of the original ROC, since in this case, the abstention ROC dominates the original ROC. Hence, to obtain maximum AUC of the abstaining classifier, when $rnr>rpr$, we should decrease $x_1$, i.e., move $x_1$ to the left. This can be implemented in Fig. \ref{f0}, where $\tan \alpha = rpr/rnr$ and $\tan \alpha' = rpr'/rnr'$. Since $\angle \alpha' < \angle \alpha$, $rpr'/rnr' < rpr/rnr < 1$. Therefore, if we move $x_2$ to $x_2'$, the difference of the two reject rates becomes larger, which violates the intention that $rnr$ and $rpr$ should be as approximate as possible. Likewise, when $rpr>rnr$, we should move $x_2$ to the right. If one of the two constraints is satisfied, lines 21-26 are enforced. In this situation, the enforcement of lines 21-26 is equivalent to the implementation of the statement in (\ref{e5}). For example, if $rnr > n_{max}$ and $rpr \leq p_{max}$, we can obtain that $rnr > n_{max} = p_{max} \geq rpr$.

When $n_{max} = p_{max} \times times$ and $times >$ 1, to make the reject rates controllable, $rnr$ and $rpr \times times$ should be approximated. Hence, when $rnr > rpr \times times$, we move $x_1$ to the left; otherwise, move $x_2$ to the right. The process stops until $rnr \leq n_{max}$. At this moment, the constraint $rpr \leq p_{max}$ is usually not satisfied, and line 25 of Algorithm 1 is performed. When $p_{max} = n_{max} \times times$ and $times >$ 1, the opposite operation is carried out. 

Once the constraints $rnr \leq n_{max}$ and $rpr \leq p_{max}$ are satisfied, the searching process is terminated. Then we calculate the reject thresholds $t_1$ and $t_2$. The BA2 classifier based on $t_1$ and $t_2$ has the optimal AUC. If $t_2$ becomes large or $t_1$ becomes small, the performance constraints are not satisfied any more. If $t_2$ becomes small or $t_1$ becomes large, the rejection interval [$t_1,t_2$] becomes small. This causes that the possible errors may not be rejected and therefore, the AUC decreases.

\section{Experimental Framework}

\subsection{Experimental Datasets and Setups}
We perform the experiments using four real-world datasets, which are available from the UCI repository~\cite{asuncion2007uci}. The characteristics of the four datasets~\cite{lin2017twin} are listed in Table~\ref{t1}. For each dataset, we perform a stratified ten-fold cross validation, of which nine folds are used to determine the reject thresholds $t_1$ and $t_2$, and the remaining fold is used to obtain the performance of the compared methods. In the process of determining the thresholds $t_1$ and $t_2$, we utilize a nine-fold cross validation to generate a more smooth ROC curve. That is, among the nine folds, we use eight folds as the training set and the remaining fold as the test set to generate a ROC curve. Then, the resulting nine ROC curves are averaged using threshold averaging method~\cite{fawcett2004roc}. The entire ten-fold cross validation is repeated ten times, and we present the averaged performance metrics.
\begin{table}[htbp]
  \centering
  \caption{Characteristics of the real-world datasets}
    \begin{tabular}{p{7em}p{6em}p{6em}p{6em}}
    \toprule[0.5pt]
    dataset & \# Pos.  & \# Neg. & \# Attr. \\
    \midrule[0.5pt]
    German credit & 300   & 700   & 20 \\
    hepatitis & 32    & 123   & 19 \\
    cmc   & 333   & 1140  & 9 \\
    abalone & 335   & 3842  & 8 \\
    \bottomrule[0.5pt]
    \end{tabular}%
  \label{t1}%
\end{table}%

In the paper, two groups of experiments are enforced to evaluate the BA2 method by comparing with bounded-abstaining  classifier (BA)~\cite{pietraszek2007use} in Section~\ref{sec3.2} and comparing with twin SVM with reject option (RO-TWSVM)~\cite{lin2017twin} in Section~\ref{sec3.3}, respectively.

\subsection{Comparison of BA2 and BA} \label{sec3.2}
BA is selected to compare with BA2, since the ideas of BA and BA2 are similar, which search the two reject thresholds using ROC curves by restricting the maximum ratios of rejected examples. The difference is that BA is realized by minimizing the misclassification cost and restricting the total reject rate less than a given value; whereas BA2 is obtained by maximizing the AUC value and controlling the proportions of the rejected positive and negative examples separately. We use $k$-NN as the scoring classifier to build the ROC curve~\cite{vanderlooy2009roc}. Considering the small size of the datasets, we set $k$ = 3. Here, we use the area under the ROC curve ($AUC$), the sensitivity ($Sen$), the rejected positive rate ($Rpr$), and the rejected negative rate ($Rnr$) as evaluation metrics.
\begin{table}[htbp]
  \centering
  \caption{Results of BA(1) ($CR$ = 1) and BA2 at 0.1/0.2/0.3-reject.}
    \begin{tabular}{p{7em}p{5em}p{5em}p{5em}p{5em}p{5em}p{5em}}
    \toprule
         & \multicolumn{2}{c}{0.1-reject} & \multicolumn{2}{c}{0.2-reject} & \multicolumn{2}{c}{0.3-reject} \\
\cmidrule(lr){2-3}\cmidrule(lr){4-5}\cmidrule(lr){6-7}
          & BA(1) & BA2   & BA(1) & BA2   & BA(1) & BA2 \\
    \midrule
  German credit &       &       &       &       &       &  \\
    $AUC$   & \textbf{0.7352 } & \textbf{0.7352 } & 0.7469  & \textbf{0.7509 } & 0.7453  & \textbf{0.7632 } \\
    $Sen$   & 0.4651  & \textbf{0.7159 } & 0.4645  & \textbf{0.7390 } & 0.4484  & \textbf{0.7595 } \\
    $Rpr$   & 0.1367  & 0.0937  & 0.2727  & 0.1917  & 0.3893  & 0.2953  \\
    $Rnr$   & 0.0920  & 0.0906  & 0.1759  & 0.1984  & 0.2627  & 0.2964  \\
    hepatitis &       &       &       &       &       &  \\
    $AUC$   & 0.8807  & \textbf{0.9245 } & 0.8859  & \textbf{0.9266 } & 0.9246  & \textbf{0.9523 } \\
    $Sen$   & 0.5476  & \textbf{0.7356 } & 0.5690  & \textbf{0.7615 } & 0.5994  & \textbf{0.8468 } \\
    $Rpr$   & 0.0717  & 0.0317  & 0.0857  & 0.0486  & 0.1155  & 0.0859  \\
    $Rnr$   & 0.0692  & 0.0851  & 0.1705  & 0.1958  & 0.3106  & 0.2878  \\
    cmc   &       &       &       &       &       &  \\
    $AUC$   & 0.6412  & \textbf{0.6581 } & 0.6098  & \textbf{0.6619 } & 0.5759  & \textbf{0.6649 } \\
    $Sen$   & 0.2563  & \textbf{0.4993 } & 0.2085  & \textbf{0.5171 } & 0.1442  & \textbf{0.5204 } \\
    $Rpr$   & 0.1513  & 0.0901  & 0.3040  & 0.2018  & 0.4216  & 0.3010  \\
    $Rnr$   & 0.0790  & 0.0885  & 0.1608  & 0.1975  & 0.2523  & 0.2910  \\
    abalone &       &       &       &       &       &  \\
    $AUC$   & 0.8469  & \textbf{0.8656 } & 0.8195  & \textbf{0.8783 } & 0.7843  & \textbf{0.8934 } \\
    $Sen$   & 0.4635  & \textbf{0.7821 } & 0.3877  & \textbf{0.8033 } & 0.2573  & \textbf{0.8291 } \\
    $Rpr$   & 0.2970  & 0.0901  & 0.5509  & 0.1862  & 0.7294  & 0.2817  \\
    $Rnr$   & 0.0842  & 0.0962  & 0.1719  & 0.2007  & 0.2658  & 0.2945  \\
    \bottomrule
    \end{tabular}%
  \label{t2}%
\end{table}%
\begin{table}[tbp]
  \centering
  \caption{Results of BA(0.5) ($CR$ = 0.5) and BA2 with different values of $p_{max}$ and $n_{max}$}
    \begin{tabular}{p{7em}p{5em}p{5em}p{5em}p{5em}p{5em}p{5em}}
    \toprule
          & \multicolumn{3}{c}{BA(0.5)} & \multicolumn{3}{c}{BA2} \\
\cmidrule(lr){2-4}\cmidrule(lr){5-7}
          & 0.1-reject & 0.2-reject & 0.3-reject & \multicolumn{1}{l}{(0.1, 0.2)} & \multicolumn{1}{l}{(0.1, 0.3)} & \multicolumn{1}{l}{(0.2,0.3)} \\
    \midrule
    German credit &       &       &       &       &       &  \\
    $AUC$   & 0.7331  & 0.7353  & 0.7554  & 0.7200  & 0.6947  & 0.7413  \\
    $Sen$   & 0.8297  & 0.8380  & 0.8266  & 0.9192  & 0.9853  & 0.8934  \\
    $Rpr$   & 0.0797  & 0.1603  & 0.2660  & 0.1040  & 0.1007  & 0.1980  \\
    $Rnr$   & 0.1076  & 0.2164  & 0.3149  & 0.2127  & 0.2847  & 0.3019  \\
    hepatitis &       &       &       &       &       &  \\
    $AUC$   & 0.8975  & 0.9416  & 0.9579  & 0.9376  & 0.9293  & 0.9529  \\
    $Sen$   & 0.7938  & 0.7739  & 0.7871  & 0.9367  & 0.9789  & 0.9500  \\
    $Rpr$   & 0.0277  & 0.0683  & 0.0865  & 0.0570  & 0.0564  & 0.0643  \\
    $Rnr$   & 0.1147  & 0.2208  & 0.3018  & 0.1966  & 0.2792  & 0.2938  \\
    cmc   &       &       &       &       &       &  \\
    $AUC$   & 0.6617  & 0.6656  & 0.6703  & 0.6503  & 0.6569  & 0.6543  \\
    $Sen$   & 0.6802  & 0.6872  & 0.6519  & 0.9128  & 0.9251  & 0.7927  \\
    $Rpr$   & 0.0822  & 0.1751  & 0.2637  & 0.0161  & 0.0048  & 0.2110  \\
    $Rnr$   & 0.1054  & 0.2114  & 0.3245  & 0.0355  & 0.0090  & 0.3375  \\
    abalone &       &       &       &       &       &  \\
    $AUC$   & 0.8656  & 0.8805  & 0.8884  & 0.8664  & 0.8598  & 0.8836  \\
    $Sen$   & 0.7780  & 0.7504  & 0.7280  & 0.9060  & 0.9769  & 0.8865  \\
    $Rpr$   & 0.1080  & 0.2566  & 0.4225  & 0.0924  & 0.1056  & 0.1850  \\
    $Rnr$   & 0.1034  & 0.1994  & 0.2924  & 0.1979  & 0.2965  & 0.3011  \\
    \bottomrule
    \end{tabular}%
  \label{t3}%
\end{table}%

Since BA only restricts the overall reject rate, to ensure comparability, we firstly set the same value for $p_{max}$ and $n_{max}$ in BA2, where $p_{max}$ and $n_{max}$ are the preset upper bounds that $rpr$ and $rnr$ should not exceed. We set the upper bounds of the reject rates at 0.1, 0.2 and 0.3, denoted as 0.1/0.2/0.3-reject. In BA, the cost is set by defining the cost ratio $CR=c_{01}/c_{10}$, where $c_{01}$ is the cost of misclassifying a negative example as positive, and $c_{10}$ is the cost of misclassifying a positive example as negative. We set $CR$ = 1, which assumes that the error costs of two classes are equal, and this setting is denoted as BA(1). The results of $AUC$, $Sen$, $Rpr$, and $Rnr$ using 0.1/0.2/0.3-reject are shown in Table~\ref{t2}.

We can observe that BA2 has larger $AUC$ and sensitivity values than BA(1) on the four real-world datasets. When the preset maximum reject rates increase, the values of the performance metrics $AUC$ and sensitivity increase in BA2; whereas in BA(1), their values decrease on datasets \textit{cmc} and \textit{abalone}. Also, in BA(1), the values of $Rpr$ and $Rnr$ are not controllable, and the values of $Rpr$ are much higher than the preset bounds on the last two datasets. By contrast, the values of $Rpr$ and $Rnr$ in BA2 can be controlled by the setting parameters $p_{max}$ and $n_{max}$. The controllable reject rates of two classes is of great significance in practical applications, since we can set acceptable $p_{max}$ and $n_{max}$ according to the actual application requirements of human and financial resources.

Considering the unequal error costs, we compare BA and BA2 by setting $CR <$ 1 in BA and setting different values for $p_{max}$ and $n_{max}$ in BA2. Usually, the positive class has higher error cost and has less examples in risk-related fields, and in the four datasets used in the paper, the number of positive examples is exactly smaller than that of negative examples. Therefore, we set $CR$ = 0.5, following the setup in~\cite{pietraszek2007use}. Likewise, because of the higher cost of the positive class, we set $p_{max} < n_{max}$. The results of BA with $CR$ = 0.5 and BA2 using different values of $p_{max}$ and $n_{max}$ are shown in Table~\ref{t3}. BA(0.5) means the setting of $CR$ = 0.5 in BA, and (0.1,0.2) means the values of $p_{max}$ and $n_{max}$ are 0.1 and 0.2 in BA2, respectively. 

Although the comparability between BA(0.5) and BA2 with different $p_{max}$ and $n_{max}$ is small, an obvious observation is that when the $AUC$ values of BA(0.5) and BA2 are similar, the sensitivity values of BA2 are much higher than that of BA(0.5). And in BA2, almost all the values of $Rpr$ and $Rnr$ are controllable when $p_{max}$ and $n_{max}$ are set using different values. This is very important when error costs are unequal and the cost of the false negative is higher than the cost of the false positive. In this case, BA2 can be set using small $p_{max}$ and large $n_{max}$, and thus high sensitivity can be obtained while good results of $AUC$ are achieved. In addition, in BA, comparing with the values of $Rpr$ with $CR = 1$, the values of $Rpr$ with $CR < 1$ decrease.

\subsection{Comparison of BA2 and RO-TWSVM} \label{sec3.3}
RO-TWSVM improves the previous SVM with reject option (RO-SVM)~\cite{tortorella2004reducing} using twin SVM instead of SVM. In RO-TWSVM, the ROC curve is built according to the scores obtained by TWSVM, and the reject thresholds are determined by minimizing the total cost. The total cost is defined as
\begin{align}  \label{e6}
cost(t_1,t_2)&=r(pos)\cdot CFN \cdot fnr(t_1)+r(neg) \cdot CTN \cdot tnr(t_1) \notag \\ 
& +r(pos)\cdot CTP \cdot tpr(t_2) + r(neg) \cdot CFP \cdot fpr(t_2) \\
& + r(pos)\cdot CR \cdot rpr(t_1,t_2) + r(neg) \cdot CR \cdot rnr(t_1,t_2) \notag 
\end{align}
where $r(pos)$ and $r(neg)$ are the ratios of the positive examples and the negative examples in the training set, respectively. $CFP$, $CFN$ and $CR$ are the costs of false positive errors, false negative errors and rejection, respectively. $CTP$ and $CTN$ are the costs of true positive and true negative, respectively. We adopt three cost models used in~\cite{lin2017twin}, which are shown in Table~\ref{t4}. Here, we do not use the cost model of CM2. In CM2, the mean cost of $CFN$ (Unif [0,50]) is lower than the mean cost of $CFP$ (Unif [0,100]). In the experiment, we assume the misclassification cost of the positive class is higher than that of the negative class. To compare with RO-TWSVM, we also use TWSVM to obtain the example scores in BA2 (BA2-TWSVM), and we utilize four same datasets (Table \ref{t5}) in \cite{lin2017twin}, which are available in KEEL dataset repository \cite{alcala2011keel}.
\begin{table}[htbp]
  \centering
  \caption{The cost models used in~\cite{lin2017twin}}
    \begin{tabular}{p{3em}p{6em}p{6em}p{6em}p{6em}p{6em}}
    \toprule[0.5pt]
    \multicolumn{1}{r}{} & $CTP$   & $CFP$   & $CTN$   & $CFN$ & $CR$ \\
    \midrule[0.5pt]
    CM1   & Unif [-10,0] & Unif [0,50] & Unif [-10,0] & Unif [0,50] & \multicolumn{1}{l}{1} \\
    CM3   & Unif [-10,0] & Unif [0,50] & Unif [-10,0] & Unif [0,100] & \multicolumn{1}{l}{1} \\
    CM4   & Unif [-10,0] & Unif [0,50] & Unif [-10,0] & Unif [0,50] & Unif [0,30] \\
    \bottomrule[0.5pt]
    \end{tabular}%
  \label{t4}%
\end{table}%

\begin{table}[htbp]
  \centering
  \caption{Characteristics of the KEEL datasets}
    \begin{tabular}{p{15em}p{5em}p{5em}p{5em}}
    \toprule
    Dataset &  \# Pos.   & \# Neg. & \# Attr. \\
    \midrule
    Pima & 268 & 500 & 8 \\
    German credit (GC) & 300 & 700 & 20 \\
    Breast cancer Wisconsin (WBC) & 239 & 460 & 9 \\
    Heart disease Cleveland (CHD) & 83  & 214 & 13 \\
    \bottomrule
    \end{tabular}%
  \label{t5}%
\end{table}%

We conduct the Wilcoxon rank sum test \cite{lin2017twin} on the four real-world datasets for the comparison of RO-TWSVM and BA2-TWSVM. In the Wilcoxon rank sum test, for each cost model in Table \ref{t4}, 1000 groups of cost terms of $CTP$, $CFP$, $CTN$, $CFN$ and $CR$ are generated. For each cost group, the total costs of two compared methods are computed according to the equation (\ref{e6}) and AUC values of the two abstaining classifiers are also calculated. Finally, the numbers of cases where values of BA2-TWSVM is higher, lower or identical than values of RO-TWSVM in terms of total cost and AUC are counted. The identical case includes two conditions that the costs or AUC values of the compared methods are equal or the cost term is not applicable to take reject option in RO-TWSVM. The details of Wilcoxon rank sum test can be found in \cite{lin2017twin}. Considering  the need of setting the parameters of reject rates in BA2, we firstly perform RO-TWSVM, and use the average values of the rejected positive rate and the rejected negative rate in each cost group as $p_{max}$ and $n_{max}$ in BA2-TWSVM, respectively. The linear kernel function is used in TWSVM. The comparison results of cost and $AUC$ are shown in Tables~\ref{t6} and \ref{t7}, respectively, where in each scenarios, the three numbers are the counts of BA2-TWSVM having higher, lower or identical value compared with RO-TWSVM.
\begin{table}[htbp]
  \centering
  \caption{Cost results of RO-TWSVM and BA2-TWSVM for linear kernel based on Wilcoxon rank sum test}
    \begin{tabular}{*{5}{p{5em}}}
    \toprule
        & Pima & GC & WBC & CHD  \\
    \midrule
    CM1  & 104 & 305 & 311& 217 \\
         & \textbf{782} & \textbf{597} & \textbf{591}& \textbf{685} \\
        & 98  & 98  & 98  & 98 \\
        \\
    CM3  & 233 & 312 & 267& 162 \\
         & \textbf{704} & \textbf{625} & \textbf{670}& \textbf{775} \\
        & 63  & 63  & 63  & 63 \\
        \\
    CM4  & 112 & 136 & 143& 161 \\
         & \textbf{452} & 428 & 421& 403 \\
         & 436 & \textbf{436} & \textbf{436}& \textbf{436} \\
    \bottomrule
    \end{tabular}%
  \label{t6}%
\end{table}%

\begin{table}[htbp]
  \centering
  \caption{AUC results of RO-TWSVM and BA2-TWSVM for linear kernel based on Wilcoxon rank sum test}
    \begin{tabular}{*{5}{p{5em}}}
    \toprule
        & Pima & GC & WBC & CHD  \\
    \midrule
    CM1   & \textbf{803} & \textbf{829} & 54& \textbf{615} \\
         & 82  & 73  & \textbf{840}& 287 \\
         & 99  & 98 & 106 & 98 \\
        \\
    CM3   & \textbf{694} & \textbf{903} & 44& \textbf{637} \\
         & 243 & 34 & \textbf{887} & 300 \\
          & 63  & 63 & 69 & 63 \\
        \\
    CM4  & 286 & 310 & 194& 229 \\
         & 238 & 203 & 248& 335 \\
         & \textbf{476} & \textbf{487} & \textbf{558}& \textbf{436} \\
    \bottomrule
    \end{tabular}%
  \label{t7}%
\end{table}%

In Table \ref{t6}, for CM1 and CM3, the number of cases that BA2-TWSVM produces lower costs than RO-TWSVM is much more than the number of the higher or identical cases. For CM4, due to the variable reject cost $CR$, the case that the cost term is not suitable for rejection becomes more. However, in the remaining two cases, the number of BA2-TWSVM's costs lower than RO-TWSVM's costs is more than the number of the opposite case. In Table \ref{t7}, except on the dataset \textit{WBC}, the number of cases that BA2-TWSVM has higher $AUC$ values than RO-TWSVM is significantly more than the number of the other two cases for CM1 and CM3. For dataset \textit{WBC}, the classifier TWSVM obtains high $AUC$ value (larger than 0.99), so the examples are easily discriminative, and thus the reject option is not essential. For other hard datasets , BA2-TWSVM performs better than RO-TWSVM in terms of total cost and $AUC$.

\section{Conclusions}
In this paper, we propose the BA2 method for abstaining classification, which avoids the introduction of the cost information. The BA2 method achieves higher AUC compared with BA and it has lower cost compared with RO-TWSVM. What's more, BA2 can control the respective reject rates of the positive and negative classes, which is extremely essential in risk-associated fields, such as medical diagnosis. Acceptable upper bounds of the reject rates can be set according to the actual application requirements of human and financial resources. In future, we would like to employ the method to specific risk-associated fields and to the field of big data with imbalance.

%
%
\bibliographystyle{splncs04}
\bibliography{samplepaper}

\end{document}